\documentclass[journal]{IEEEtran}

\usepackage{amsmath}
\usepackage{graphics, theorem, times, amsfonts, amsmath, amssymb, cite, caption, graphicx, array}
\usepackage{tikz}
\usetikzlibrary{shapes,arrows}
\usepackage{pgfplots}
\usepackage{color}
\usepackage{needspace}

\usepackage{hyperref}
\usepackage{multirow}
\usepackage{rotating}
\usepackage{flushend, cleveref, subcaption}
\input{mysymbol.sty}

\newtheorem{example}{\hspace{0pt}\bf Example}

\newtheorem{remark}{\hspace{0pt}\bf Remark}

\newcolumntype{x}[1]{%
>{\centering\hspace{0pt}}p{#1}}%
\begin{document}
\title{Authorship Attribution through Function\\ Word Adjacency Networks}
\author{Santiago Segarra, Mark Eisen, and Alejandro Ribeiro 
\thanks{Supported by NSF CAREER CCF-0952867 and NSF CCF-1217963. The authors are with the Department of Electrical and Systems Engineering, University of Pennsylvania, 200 South 33rd Street, Philadelphia, PA 19104. Email: \{ssegarra, maeisen, aribeiro\}@seas.upenn.edu. Part of the results in this paper appeared in \cite{cSegarraetal13}.}}

\maketitle

\begin{abstract}
A method for authorship attribution based on function word adjacency networks (WANs) is introduced. Function words are parts of speech that express grammatical relationships between other words but do not carry lexical meaning on their own. In the WANs in this paper, nodes are function words and directed edges stand in for the likelihood of finding the sink word in the ordered vicinity of the source word. WANs of different authors can be interpreted as transition probabilities of a Markov chain and are therefore compared in terms of their relative entropies. Optimal selection of WAN parameters is studied and attribution accuracy is benchmarked across a diverse pool of authors and varying text lengths. This analysis shows that, since function words are independent of content, their use tends to be specific to an author and that the relational data captured by function WANs is a good summary of stylometric fingerprints. Attribution accuracy is observed to exceed the one achieved by methods that rely on word frequencies alone. Further combining WANs with methods that rely on word frequencies alone, results in larger attribution accuracy, indicating that both sources of information encode different aspects of authorial styles.
\end{abstract}


\section{Introduction}\label{sec:introduction}

The discipline of authorship attribution is concerned with matching a text of unknown or disputed authorship to one of a group of potential candidates. More generally, it can be seen as a way of quantifying literary style or uncovering a stylometric fingerprint. The most traditional application of authorship attribution is literary research, but it has also been applied in forensics \cite{Grant07}, defense intelligence \cite{AbbasiChen05} and plagiarism \cite{Deckeretal07}. Both, the availability of electronic texts and advances in computational power and information processing, have boosted accuracy and interest in computer based authorship attribution methods \cite{Holmes94, Juola06, Stamatatos09}.

Authorship attribution dates at least to more than a century ago with a work that proposed distinguishing authors by looking at word lengths \cite{Mendenhall87}. This was later improved by \cite{Yule39} where the average length of sentences was considered as a determinant. A seminal development was the introduction of the analysis of function words to characterize authors' styles \cite{MostWall64} which inspired the development of several methods. Function words are words like prepositions, conjunctions, and pronouns which on their own carry little meaning but dictate the grammatical relationships between words. The advantage of function words is that they are content independent and, thus, can carry information about the author that is not biased by the topic of the text being analyzed. Since \cite{MostWall64}, function words appeared in a number of papers where the analysis of the frequency with which different words appear in a text plays a central role one way or another; see e.g., \cite{Burrows89, HolFors95, Hoover04, Halterenetal05}. Other attribution methods include the stylometric techniques in \cite{ForsHol96}, the use of vocabulary richness as a stylometric marker \cite{Yule44, Holmes91, TweBaa98} -- see also \cite{Hoover03} for a critique --, the use of stable words defined as those that can be replaced by an equivalent \cite{Koppel06}, and syntactical markers such as taggers of parts of speech \cite{Cutting92}. 

In this paper, we use function words to build stylometric fingerprints but, instead of focusing on their frequency of usage, we consider their relational structure. We encode these structures as word adjacency networks (WANs) which are asymmetric networks that store information of co-appearance of two function words in the same sentence (Section \ref{sec:wan_method}). With proper normalization, edges of these networks describe the likelihood that a particular function word is encountered in the text given that we encountered another one. In turn, this implies that WANs can be reinterpreted as Markov chains describing transition probabilities between function words. Given this interpretation it is natural to measure the dissimilarity between different texts in terms of the relative entropy between the associated Markov chains (Section \ref{sec:network_similarity}). Markov chains have also been used as a tool for authorship attribution in \cite{Khemelev01, Kuku01}. However, the chains in these works represent transitions between letters, not words. Although there is little intuitive reasoning behind the notion that an author's style can be modeled by his usage of individual letters, these approaches generate somewhat positive results.

The classification accuracy of WANs depends on various parameters regarding the generation of the WANs as well as the selection of words chosen as network nodes. We consider the optimal selection of these parameters and develop an adaptive strategy to pick the best network node set given the texts to attribute (Section \ref{sec:optimal_number_of_function_words}). Using a corpus composed of texts by 21 authors from the 19th century, we illustrate the implementation of our method and analyze the changes in accuracy when modifying the number of candidate authors as well as the length of the text of known (Section \ref{sec:varying_profile_length}) and unknown (Section \ref{sec:varying_text_length}) authorship. Further, we analyze how the similarity of styles between two authors influences the accuracy when distinguishing their texts (Section \ref{sec:inter_profile_distance}). We then incorporate authors from the early 17th century to the corpus and analyze how differences in time period, genre, and gender influence the classification rate of WANs (Sections \ref{sec:time} to \ref{sec:gender}). We also show that WANs can be used to detect collaboration between several authors (Section \ref{sec:detecting_collaborations}). We further demonstrate that our classifier performs better than techniques based on function word frequencies alone (Section \ref{sec:comparison_combination_frequency_based}). Perhaps more important, we show that the stylometric information captured by WANs is not the same as the information captured by word frequencies. Consequently, their combination results in a further increase in classification accuracy.

\section{Problem Formulation} \label{sec:problem_formulation}

We are given a set of $n$ authors $A=\{a_1, a_2, ... , a_n\}$, a set of $m$ known texts $T=\{t_1, t_2, ... , t_m\}$ and a set of $k$ unknown texts $U=\{u_1, u_2, ... , u_k\}$. We are also given an authorship attribution function $r_T:T \to A$ mapping every known text in $T$ to its corresponding author in $A$, i.e. $r_T(t) \in A$ is the author of text $t$ for all $t \in T$. We further assume $r_T$ to be surjective, this implies that for every author $a_i\in A$ there is at least one text $t_j\in T$ with $r_T(t_j)=a_i$. Denote as $T^{(i)} \subset T$ the subset of known texts written by author $a_i$, i.e.
\begin{equation}\label{eqn_def_T_i}
T^{(i)}=\{ t \,\, | \,\, t \in T, r_T(t)=a_i\}.
\end{equation}
According to the above discussion, it must be that $|T^{(i)}|>0$ for all $i$ and $\{ T^{(i)}\}^n_{i=1}$ must be a partition of $T$. In Section \ref{sec:wan_method}, we use the texts contained in $T^{(i)}$ to generate a relational profile for author $a_i$. There exists an unknown attribution function $r_{U}: U \to A$ which assigns each text $u \in U$ to its actual author $r_{U}(u)\in A$. Our objective is to approximate this unknown function with an estimator $\hat{r}_{U}$ built with the information provided by the attribution function $r_T$.  In particular, we construct word adjacency networks (WANs) for the known texts $t\in T$  and unknown texts $u\in U$. We attribute texts by comparing the WANs of the unknown texts $u\in U$ to the WANs of the known texts $t\in T$ .

In constructing WANs, the concepts of sentence, proximity, and function words are important. Every text consists of a sequence of sentences, where a sentence is defined as an indexed sequence of words between two stopper symbols. We think of these symbols as grammatical sentence delimiters,  but this is not required. For a given sentence, we define a directed proximity between two words parametric on a discount factor $\alpha \in (0,1)$ and a window length $D$. If we denote as $i(\omega)$ the position of word $\omega$ within its sentence the directed proximity $d(\omega_1, \omega_2)$ from word $\omega_1$ to word $\omega_2$ when $0< i(\omega_2) - i(\omega_1) \leq D$ is defined as 
\begin{equation}\label{eqn_def_prximity}
    d(\omega_1, \omega_2) := \alpha^{i(\omega_2) - i(\omega_1)-1}.
\end{equation}
In every sentence there are two kind of words: function and non-function words \cite{Comprehensive_Grammar}. While in \eqref{eqn_def_prximity} the words $w_1$ and $w_2$ need not be function words, in this paper we are interested only in the case in which both  $w_1$ and $w_2$ are function words. Function words are words that express primarily a grammatical relationship. These words include conjunctions (e.g., \emph{and, or}), prepositions (e.g., \emph{in, at}), quantifiers (e.g., \emph{some, all}), modals (e.g., \emph{may, could}), and determiners (e.g., \emph{the, that}). We  exclude gender specific pronouns (\emph{he, she}) as well as pronouns that depend on narration type (\emph{I, you}) from the set of function words to avoid biased similarity between texts written using the same grammatical person -- see Section \ref{sec:wan_method} for details. The 30 function words that appear most often in our experiments are listed in Table \ref{tab_function_words}. The concepts of sentence, proximity, and function words are illustrated in the following example.

\begin{table}[t]
\centering
\begin{tabular}{| c | c | c | c | c | c | c | c | c | c |}
\hline
\multicolumn{10}{|c|}{Common Function Words} \\\hline
\hline
the & and & a  & of & to & in  & that & with & as & it \\
\hline
for & but & at  & on & this & all  & by & which & they & so \\
\hline
from & no & or  & one & what & if  & an & would & when & will \\
\hline
\end{tabular}
\caption{Most common function words in analyzed texts.}
\label{tab_function_words}
\end{table}

\begin{example}\label{ex_10} \normalfont Define the set of stopper symbols as $\{. \,\,; \}$, let  the parameter $\alpha=0.8$, the window $D=4$, and consider the text
\begin{itemize} \item[] ``A swarm in May is worth a load of hay; a swarm in June is worth a silver spoon; but a swarm in July is not worth a fly.'' \end{itemize}
The text is composed of three sentences separated by the delimiter  $\{\ ; \}$. We then divide the text into its three constituent sentences and highlight the function words
\begin{itemize}
\item[] {\bf a} swarm {\bf in} May is worth {\bf a} load {\bf of} hay 
\item[] {\bf a} swarm {\bf in} June is worth {\bf a} silver spoon
\item[] {\bf but a} swarm {\bf in} July is not worth {\bf a} fly  
\end{itemize}
The directed proximity from the first \emph{a} to \emph{swarm} in the first sentence is $\alpha^0=1$ and the directed proximity from the first \emph{a} to \emph{in} is  $\alpha^1=0.8$. The directed proximity to \emph{worth} or \emph{load} is $0$ because the indices of these words differ in more than $D=4$. \end{example}

Define the classification accuracy as the fraction of unknown texts that are correctly attributed. With $\mathbb{I}$ denoting the indicator function we can write the classification accuracy $\rho$ as
\begin{equation}\label{eqn_def_accuracy}
   \rho(\hat{r}_{U})
        =   \frac{1}{k} \sum_{u \in U} \mathbb{I}\left\{\hat{r}_{U}(u)=r_{U}(u)\right\}.
\end{equation}
We use $\rho(\hat{r}_{U})$ to gauge performance of the classifier in Sections \ref{sec:optimal_number_of_function_words} to \ref{sec:comparison_combination_frequency_based}. 


\section{Function Words Adjacency Networks} \label{sec:wan_method}

As relational structures we construct WANs for each text. These weighted and directed networks contain function words as nodes. The weight of a given edge represents the likelihood of finding the words connected by this edge close to each other in the text. Formally, from a given text $t$ we construct the network $W_{t}=(F, Q_{t})$ where $F=\{f_1, f_2, ... , f_f\}$ is the set of nodes composed by a collection of function words common to all WANs being compared and $Q_{t}: F \times F \to \reals_+$ is a similarity measure between pairs of nodes. Methods to select the elements of the node set $F$ are discussed in Section \ref{sec:optimal_number_of_function_words}.

In order to calculate the similarity function $Q_{t}$, we first divide the text $t$ into sentences $s^h_{t}$ where $h$ ranges from 1 to the total number of sentences. We denote by $s^h_{t}(e)$ the word in the $e$-th position within sentence $h$ of text $t$. In this way, we define
\begin{equation}\label{eqn_def_Q_t_i}
Q_{t} (f_i, f_j) = \sum_{h,e} \mathbb{I}\{s^h_{t}(e)=f_i\} \sum^D_{d=1} \alpha^{d-1} \,\, \mathbb{I}\{s^h_{t}(e+d)=f_j\},
\end{equation}
for all $f_i, f_j \in F$, where $\alpha \in (0,1)$ is the discount factor that decreases the assigned weight as the words are found further apart from each other and $D$ is the window limit to consider that two words are related. The similarity measure in \eqref{eqn_def_Q_t_i} is the sum of the directed proximities from $f_i$ to $f_j$ defined in \eqref{eqn_def_prximity} for all appearances of $f_i$ when the words are found at most $D$ positions apart in the same sentence. Since in general $Q_{t} (f_i, f_j)\neq Q_{t} (f_j, f_i)$, the WANs generated are directed.

\begin{example}\label{ex_20}\normalfont
Consider the same text and parameters of Example \ref{ex_10}. There are four function words yielding the set $F=\{\text{a, in, of, but}\}$. The matrix representation of the similarity function $Q_t$ is
\begin{equation}
Q_t = \bordermatrix{~ & \text{a} & \text{in} & \text{of} & \text{but} \cr
            \text{a} & 0 & 3 \times 0.8^1  & 0.8^1 & 0 \cr
            \text{in} & 2 \times 0.8^3 & 0 & 0 & 0 \cr
            \text{of} & 0 & 0 & 0 & 0 \cr
            \text{but} & 1 & 0.8^2 & 0 & 0 \cr}.
\end{equation}
The total similarity value from \emph{a} to \emph{in} is obtained by summing up the three $0.8^1$ proximity values that appear in each sentence. Although the word \emph{a} appears twice in every sentence, $Q(\text{a}, \text{a})=0$ because its appearances are more than $D=4$ words apart.
\end{example}

Using text WANs, we generate a network $W_c$ for every author $a_c \in A$ as $W_c=(F,Q_c)$ where
\begin{equation}\label{eqn_def_Q_i}
Q_c= \sum_{t \in T^{(c)}} Q_{t}.
\end{equation}
Similarities in $Q_c$ depend on the amount and length of the texts written by author $a_c$. This is undesirable since we want to be able to compare relational structures among different authors. Hence, we normalize the similarity measures as 
\begin{equation}\label{eqn_def_P_i}
\hat{Q}_c(f_i, f_j) = \frac{Q_c(f_i,f_j)}{\sum_{j} Q_c(f_i,f_j)},
\end{equation}
for all $f_i, f_j \in F$. In this way, we achieve normalized networks $\hat{P}_c=(F, \hat{Q}_c)$ for each author $a_c$. In \eqref{eqn_def_P_i} we assume that there is at least one positively weighted edge out of every node $f_i$ so that we are not dividing by zero. If this is not the case for some function word $f_i$, we fix $\hat{Q}_c(f_i, f_j) = 1/|F|$ for all $f_j$.

\begin{example}\label{ex_30}\normalfont
By applying normalization \eqref{eqn_def_P_i} to the similarity function in Example \ref{ex_20}, we obtain the following normalized similarity matrix
\begin{equation}
\hat{Q}_t = \bordermatrix{~ & \text{a} & \text{in} & \text{of} & \text{but} \cr
            \text{a} & 0 & 0.75  & 0.25 & 0 \cr
            \text{in} & 1 & 0 & 0 & 0 \cr
            \text{of} & 0.25 & 0.25 & 0.25 & 0.25 \cr
            \text{bu}t & 0.61 & 0.39 & 0 & 0 \cr}.
\end{equation}
Similarity $\hat{Q}_t$ no longer depends on the length of the text $t$ but on the relative frequency of the co-appearances of function words in the text.
\end{example}

Our claim is that every author $a_c$ has an inherent relational structure $P_c$ that serves as an authorial fingerprint and can be used towards the solution of authorship attribution problems. $\hat{P}_c=(F, \hat{Q}_c)$ estimates $P_c$ with the available known texts written by author $a_c$.

\subsection{Network Similarity} \label{sec:network_similarity}
The normalized networks $\hat{P}_c$ can be interpreted as discrete time Markov chains (MC) since the similarities out of every node sum up to 1. Thus, the normalized similarity between words $f_i$ and $f_j$ is a measure of the probability of finding $f_j$ in the words following an encounter of $f_i$.
In a similar manner, we can build a MC $P_{u}$ for each unknown text $u \in U$.

Since every MC has the same state space $F$, we use the relative entropy $H(P_1, P_2)$ as a dissimilarity measure between the chains $P_1$ and $P_2$. The relative entropy is given by
\begin{equation}\label{eqn_def_H}
H(P_1, P_2)= \sum_{i,j} \pi(f_i) P_1(f_i, f_j) \log \frac{P_1(f_i, f_j)}{P_2(f_i, f_j)},
\end{equation}
where $\pi$ is the limiting distribution on $P_1$ and we consider $0 \, \log 0$ to be equal to $0$. The choice of $H$ as a measure of dissimilarity is not arbitrary. In fact, if we denote as $w_1$ a realization of the MC $P_1$, $H(P_1, P_2)$ is proportional to the logarithm of the ratio between the probability that $w_1$ is a realization of $P_1$ and the probability that $w_1$ is a realization of $P_2$. In particular, when $H(P_1, P_2)$ is null, the ratio is 1 meaning that a given realization of $P_1$ has the same probability of being observed in both MCs \cite{Kesidis93}. We point out that relative entropy measures have also been used to compare vectors with function word frequencies \cite{Zhaoetal06}. This is unrelated to their use here as measures of the relational information captured in function WANs.

Using \eqref{eqn_def_H}, we generate the attribution function $\hat{r}_{U}(u)$ by assigning the text $u$ to the author with the most similar relational structure
\begin{equation}\label{eqn_def_r_hat_T_0}
\hat{r}_{U}(u)= a_p, \,\,\, \text{where} \,\,\, p=\argmin_c H(P_{u}, \hat{P}_c).
\end{equation}
Whenever a transition between words appears in an unknown text but not in a profile, the relative entropy in \eqref{eqn_def_r_hat_T_0} takes an infinite value for the corresponding author. In practice we compute the relative entropy in \eqref{eqn_def_H} by summing over the non-zero transitions in the profiles, 
\begin{equation}\label{eqn_def_H_in_practice}
H(P_1, P_2)= \sum_{i,j | P_2(f_i, f_j) \neq 0 } \pi(f_i) P_1(f_i, f_j) \log \frac{P_1(f_i, f_j)}{P_2(f_i, f_j)}.
\end{equation}
Observe that if there is a transition between words that appears often in the text $P_1$ but never in the profile $P_2$, the expression in \eqref{eqn_def_H_in_practice} skips the relative entropy summand. This is undesirable because the often appearance of this transition in the text network $P_1$ is a strong indication that this text was not written by the author whose profile network is $P_2$. The expression in \eqref{eqn_def_H} would capture this difference by producing an infinite value for the relative entropy. However, this infinite value is still produced if a transition between words does not appear in the author profile $P_2$ and appears just once in the text $P_1$. In this case, the null contribution to the relative entropy in \eqref{eqn_def_H_in_practice} is more reasonable than the infinity contribution in \eqref{eqn_def_H} because the rarity of the transition in both texts is an indication that the text and the profile belong to the same author. Our experiments show that the latter situation is more common than the former. Transitions rare enough so as not to appear in a profile are, for the most part, also infrequent in all texts. This is reasonable because rare combinations of function words are properties of the language more than of individual authors. We have also explored the use of Laplace smoothing to avoid infinite entropies -- see e.g., \cite[Chapter 13]{Manning08}, but \eqref{eqn_def_H_in_practice} still achieves best results in practice. 

We proceed to specify the selection of function words in $F$ as well as the choice of the parameters $\alpha$ and $D$ after the following remark.

\begin{remark}\normalfont
For the relative entropies in \eqref{eqn_def_r_hat_T_0} to be well defined, the MCs $P_u$ associated with the unknown texts have to be ergodic to ensure that the limiting distributions $\pi$ in \eqref{eqn_def_H} and \eqref{eqn_def_H_in_practice} are unique. This is true if the texts that generated $P_u$ are sufficiently long. If this is not true for a particular network, we replace $\pi(f_i)$ with the expected fraction of time a randomly initialized walk spends in state $f_i$. The random initial function word is drawn from a distribution given by the word frequencies in the text. 
\end{remark}

\begin{figure*}
        \centering
        \begin{subfigure}[b]{0.5\textwidth}
          \includegraphics[width=\textwidth]{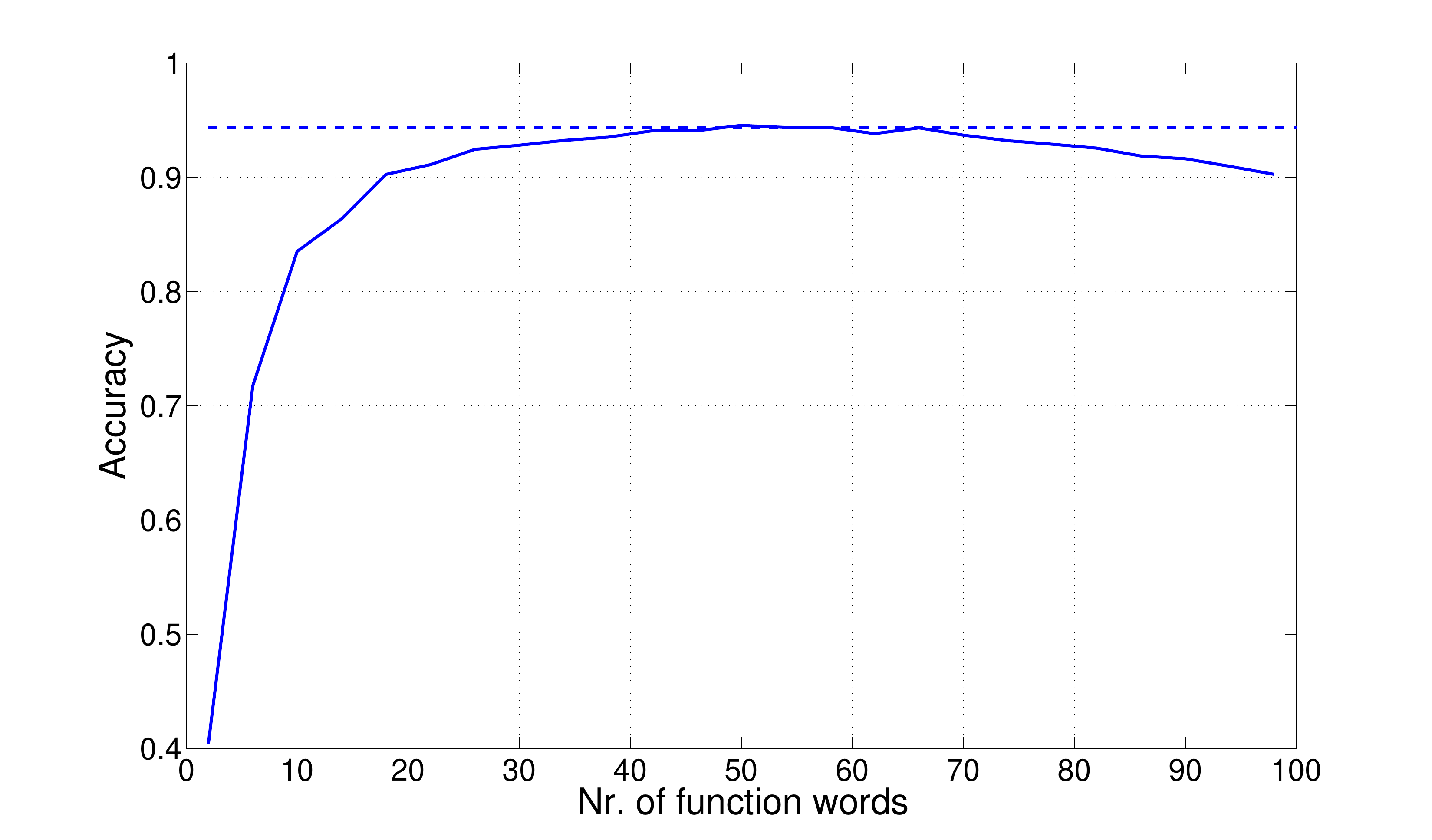}
          \caption{Attribution accuracy as a function of the network size.}
          \label{fig:adaptive_accuracy}
        \end{subfigure}%
        \begin{subfigure}[b]{0.5\textwidth}
          \includegraphics[width=\textwidth, height=0.59\textwidth]{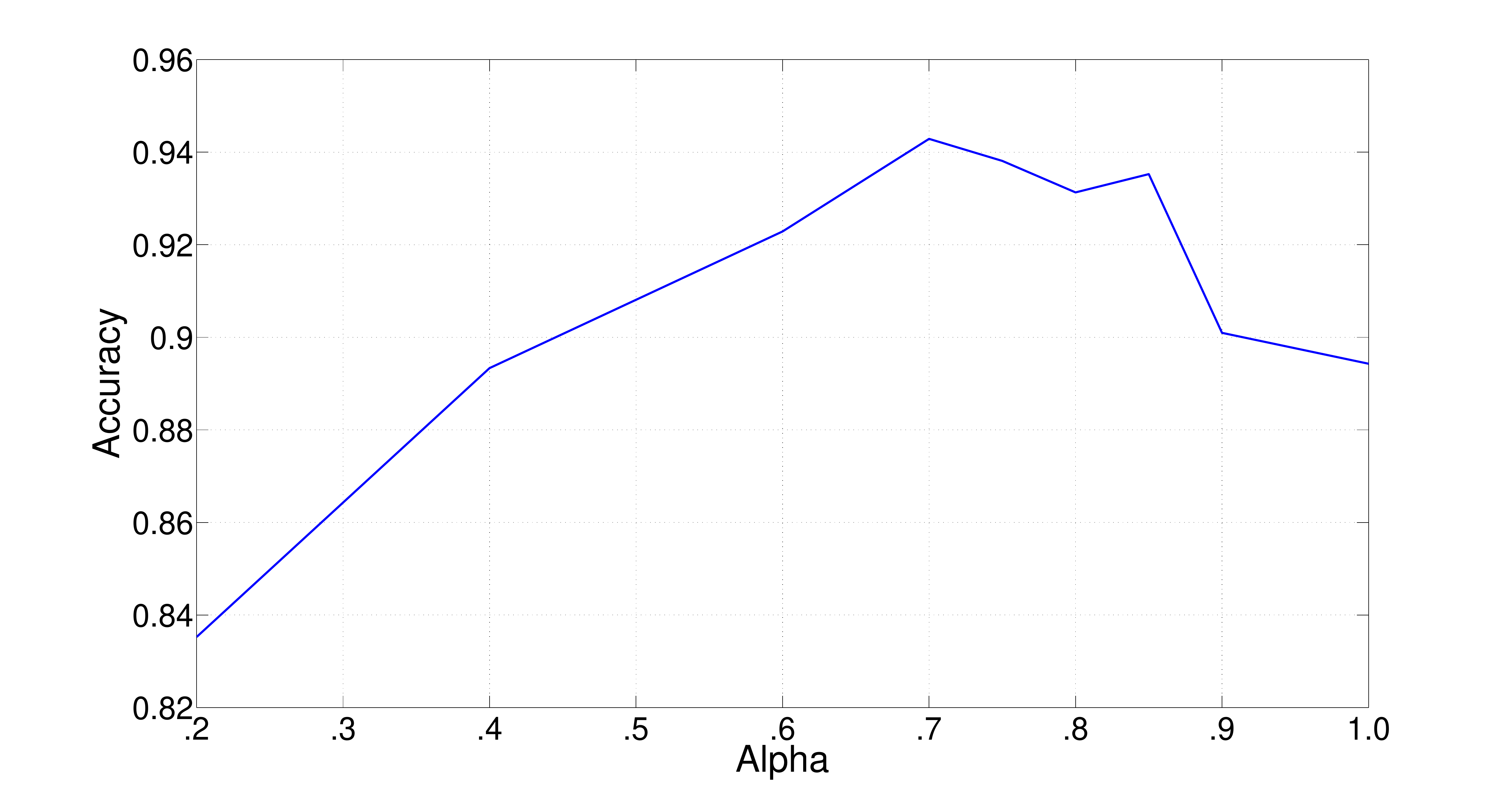}
          \caption{Attribution accuracy as a function of the discount factor $\alpha$.}
          \label{fig:best_alpha}
        \end{subfigure}
        \caption{Both figures present the accuracy for the attribution of 1,000 texts of length 10,000 words among 7 authors chosen at random with 100,000 words profiles. (a) The solid line represents the accuracy achieved for static networks of increasing size. The dashed line is the accuracy obtained by the adaptive method. (b) Accuracy is maximized for values of the discount factor $\alpha$ in the range between 0.70 and 0.85.}
\label{fig:mds}
\end{figure*}

%
\section{Selection of Function Words and WAN Parameters} \label{sec:optimal_number_of_function_words}

The classification accuracy of the function WANs introduced in Section \ref{sec:wan_method} depends on the choice of several variables and parameters: the set of sentence delimiters or stopper symbols, the window length $D$, the discount factor $\alpha$, and the set of function words $F$ defining the nodes of the adjacency networks. In this section, we study the selection of these parameters to maximize classification accuracy.

The selections of stopper symbols and window lengths are not critical. As stoppers we include the grammatical sentence delimiters `.', `?' and `!', as well as semicolons `;' to form the stopper set $\{ \text{. ? ! ;} \}$. We include semicolons since they are used primarily to connect two independent clauses \cite{Comprehensive_Grammar}. In any event, the inclusion or not of the semicolon as a stopper symbol entails a minor change in the generation of WANs due to its infrequent use. As window length we pick $D=10$, i.e., we consider that two words are not related if they appear more than 10 positions apart from each other. Larger values of $D$ lead to higher computational complexity without increase in accuracy since grammatical relations of words more than 10 positions apart are rare.

In order to choose which function words to include when generating the WANs we present two different approaches: a static methodology and an adaptive strategy. The static approach consists in picking the function words most frequently used in the union of {\it all} the texts being considered in the attribution, i.e, all those that we use to build the profile and those being attributed. By using the most frequent function words we base the attribution on repeated grammatical structures and limit the influence of noise introduced by unusual sequences of words which are not consistent stylometric markers. In our experiments, we see that selecting a number of functions words between 40 and 70 yields optimal accuracy. For way of illustration, we consider in Fig. \ref{fig:adaptive_accuracy} the attribution of 1,000 texts of length 10,000 words among 7 authors chosen at random from our pool of 19th century authors \cite{Corpus_texts_2} for a fixed value of $\alpha=0.75$ and profiles of 100,000 words -- see also Section \ref{sec:performance_analysis} for a description of the corpus. The solid line in this figure represents the accuracy achieved when using a network composed of the $n$ most common function words in the texts analyzed for $n$ going from 2 to 100. Accuracy is maximal when we use exactly 50 function words, but the differences are minimal and likely due to random variations for values of $n$ between $n=42$ and $n=66$. The flatness of the accuracy curve is convenient because it shows that the selection of $n$ is not that critical. In this particular example we can choose any value between, say $n=45$ and $n=60$, without affecting reliability. In a larger test where we also vary the length of the profiles, the length of the texts attributed, and the number of candidate authors, we find that including 60 function words is empirically optimal.

The adaptive approach still uses the most common function words but adapts the number of function words used to the specific attribution problem. In order to choose the number of function words, we implement repeated leave-one-out cross validation as follows. For every candidate author $a_i \in A$, we concatenate all the known texts $T^{(i)}$ written by $a_i$ and then break up this collection into $N$ pieces of equal length. We build a profile for each author by randomly picking $N-1$ pieces for each of them. We then attribute the unused pieces between the authors utilizing WANs of $n$ function words for $n$ varying in a given interval $[n_{\min}, n_{\max}]$. We perform $M$ of these cross validation rounds in which we change the random selection of the $N-1$ texts that build the profiles. The value of $n$ that maximizes accuracy across these $M$ trials is selected as the number of nodes for the WANs. We perform attributions using the corresponding $n$ word WANs for the profiles as well as for the texts to be attributed. In our numerical experiments we have found that using $N=10$, $n_{\min}=20$, $n_{\max}=80$, and $M$ varying between 10 and 100 depending on the available computation time are sufficient to find values of $n$ that yield good performance.

The dashed line in Fig. \ref{fig:adaptive_accuracy} represents the accuracy obtained by implementing the adaptive strategy with $N=10$, $n_{\min}=20$, $n_{\max}=80$, and $M=100$ for the same attribution problem considered in the static method -- i.e., attribution of 1,000 texts of length 10,000 words among 7 authors for $\alpha=0.75$ and profiles of 100,000 words. The accuracy is very similar to the best correct classification rate achieved by the static method. This is not just true of this particular example but also true in general. The static approach is faster because it requires no online training to select the number of words $n$ to use in the WANs. The adaptive strategy is suitable for a wider range of problems because it contains less assumptions than the static method about the best structure to differentiate between the candidate authors. E.g., when shorter texts are analyzed, experiments show that the optimal static method uses slightly less than 60 words. Likewise, the optimal choice of the number of words in the WANs changes slightly with the time period of the authors, the specific authors considered, and the choice of parameter $\alpha$. These changes are captured by the adaptive approach. We advocate adaptation in general and reserve the static method for rapid attribution of texts or cases when the number of texts available to build profiles is too small for effective cross-validation.

To select the decay parameter we use the adaptive leave-one-out cross validation method for different values of $\alpha$ and study the variation of the correct classification rate as $\alpha$ varies. In Fig. \ref{fig:best_alpha} we show the variation of the correct classification rate with $\alpha$ when attributing 1,000 texts of length 10,000 words between 7 authors of the 19th century picked at random from our text corpus \cite{Corpus_texts_2} using profiles with 100,000 words -- see also Section \ref{sec:performance_analysis} for a description of the corpus. As in the case of the number of words used in the WANs there is a wide range of values for which variations are minimal and likely due to randomness. This range lies approximately between $\alpha=0.7$ and $\alpha=0.85$. In a larger test where we also vary text and profile lengths as well as the number of candidate authors we find that $\alpha = 0.75$ is optimal. We found no gains in an adaptive method to choose $\alpha.$

\begin{figure*}
        \centering
        \begin{subfigure}[b]{0.5\textwidth}
          \includegraphics[width=\textwidth]{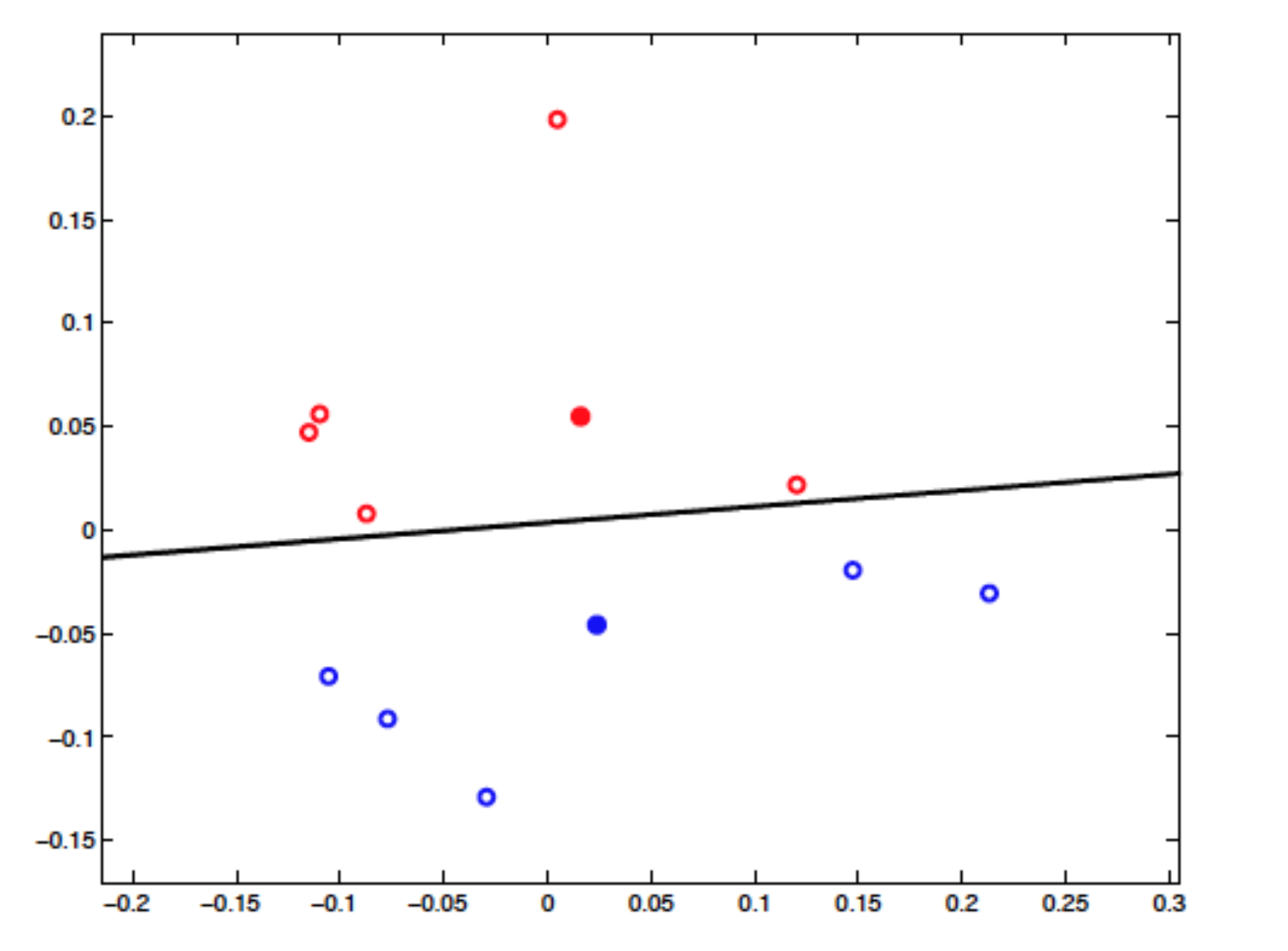}
          \caption{MDS representation for two authors.}
          \label{fig:mds_2_authors}
        \end{subfigure}%
        \begin{subfigure}[b]{0.48\textwidth}
          \includegraphics[width=\textwidth]{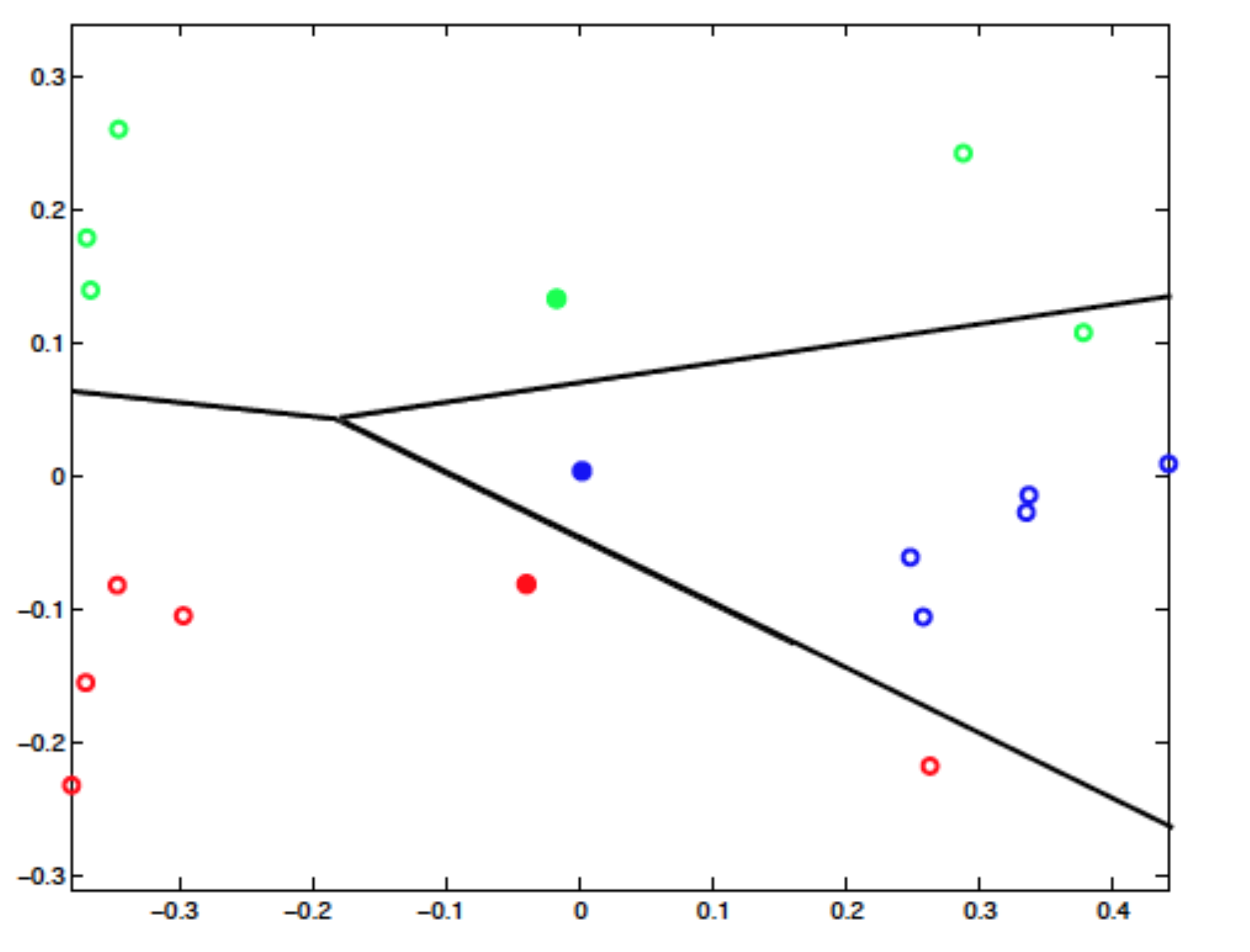}
          \caption{MDS representation for three authors. }
          \label{fig:mds_3_authors}
        \end{subfigure}
        \caption{(a) Perfect accuracy is attained for two candidate authors. Every empty circle falls in the half plane corresponding to the filled circle of their color. (b) One mistake is made for three authors. One green empty circle falls in the region attributable to the blue author.}
        \label{fig:mds}
\end{figure*}


\section{Attribution Accuracy}\label{sec:performance_analysis}

Henceforth, we fix the WAN generation parameters to the optimal values found in Section \ref{sec:optimal_number_of_function_words}, i.e., the set of sentence delimiters is \{ . ? ! ; \}, the discount factor is $\alpha=0.75$, and the window length is $D=10$. The set of function words $F$ is picked adaptively for every attribution problem by performing $M=10$ cross validation rounds.

The text corpus used for the simulations consists of authors from two different periods  \cite{Corpus_texts_2}. The first group corresponds to 21 authors spanning the 19th century, both American -- such as Nathaniel Hawthorne and Herman Melville -- and British -- such as Jane Austen and Charles Dickens. For these 21 authors, we have an average of 6.5 books per author with a minimum of 4 books for Charlotte Bronte and a maximum of 10 books for Herman Melville and Mark Twain. In terms of words, this translates into an average of 560,000 words available per author with a minimum of 284,000 words for Louisa May Alcott and a maximum of 1,096,000 for Mark Twain. The second group of authors corresponds to 7 Early Modern English playwrights spanning the late 16th century and the early 17th century, namely William Shakespeare, George Chapman, John Fletcher, Ben Jonson, Christopher Marlowe, Thomas Middleton, and George Peele. For these authors we have an average of 22 plays per author with a minimum of 4 plays for Peele and a maximum of 47 plays written either completely or partially by Fletcher. In terms of word length, we count with an average length of 400,000 words per author with a minimum of 50,000 for Peele and a maximum of 900,000 for Fletcher.

To illustrate authorship attribution with function WANs, we solve an authorship attribution problem with two candidate authors: Mark Twain and Herman Melville. For each candidate author we are given five known texts and are asked to attribute ten unknown texts, five of which were written by Twain while the other five belong to Melville \cite{Corpus_texts_2}. Every text in this attribution belongs to a different book and corresponds to a 10,000 word extract, i.e. around 25 pages of a paper back midsize edition. The five known texts from each author are used to generate corresponding profiles as described in Section \ref{sec:wan_method}. Relative entropies in \eqref{eqn_def_H_in_practice} from each of the ten unknown texts to each of the two resulting profiles are then computed. 

Since relative entropies are not metrics, we use multidimensional scaling (MDS) \cite{CoxCox08} to embed the two profiles and the ten unknown texts in 2-dimensional Euclidean metric space with minimum distortion. The result is illustrated in Fig. \ref{fig:mds_2_authors}.  Twain's and Melville's profiles are depicted as red and blue filled circles, respectively. Unknown texts are depicted as empty circles, where the color indicates the real author, i.e. red for Twain and blue for Melville. A solid black line composed of points equidistant to both profiles is also plotted. This line delimits the two half planes that result in attribution to one author or the other. From Fig. \ref{fig:mds_2_authors}, we see that the attribution is perfect for these two authors. All red (Twain) empty circles fall in the half plane closer to the filled red circle and all blue (Melville) empty circles fall in the half plane closer to the filled blue circle. We emphasize that the WAN attributions are not based on these Euclidean distances but on the non-metric dissimilarities given by the relative entropies. Since the number of points is small, the MDS distortion is minor and the distances in Fig. \ref{fig:mds_2_authors} are close to the relative entropies. The latter separate the points better, i.e., relative entropies are smaller for texts of the same author and larger for texts of different authors. 

We also illustrate an attribution between three authors by creating a profile for Jane Austen using five 10,000 word excerpts and adding five 10,000 word excerpts of texts written by Jane Austen to the ten excerpts to attribute from Twain and Melville's books. We then perform an attribution of the 15 texts to the three profiles constructed. An MDS approximate representation of the relative entropies between texts and profiles is shown in Fig. \ref{fig:mds_3_authors} where filled circles represent profiles and empty circles represent texts. Different colors are used to distinguish Twain (red), Melville (blue), and Austen (green). We also plot the Voronoi tessellation induced by the three profiles, which specify the regions of the plane that are attributable to each author. Different from the case in Fig. \ref{fig:mds_2_authors}, attribution is not perfect since one of Austen's texts is mistakenly attributed to Melville. This is represented in Fig \ref{fig:mds_3_authors} by the green empty circle that appears in the section of the Voronoi tessellation that corresponds to the blue profile. 

Besides the number of authors, the other principal determinants of classification accuracy are the length of the profiles, the length of the texts of unknown authorship, and the similarity of writing styles as captured by the relative entropy dissimilarities between profiles. We study these effects in sections \ref{sec:varying_profile_length},\ref{sec:varying_text_length}, and \ref{sec:inter_profile_distance}, respectively.


\subsection{Varying Profile Length}\label{sec:varying_profile_length}

\begin{table*}
\centering
\begin{tabular}{|c|l|l|l|l|l|l|l|l|l|l|l|l|l|l|l|l|l|}
\hline
\multirow{2}{*}{Nr. of authors} &\multicolumn{10}{|c|}{Number of words in profile (thousands)} & \multirow{2}{*}{Rand.} \\
\cline{2-11}
&\textbf{10}&\textbf{20}&\textbf{30}&\textbf{40}&\textbf{50}&\textbf{60}&\textbf{70}&\textbf{80}&\textbf{90}&\textbf{100}&\\\hline
\textbf{2}&0.927&0.964&0.984&0.985&0.981&0.979&0.981&0.986&0.992&0.988&0.500\\\hline
\textbf{3}&0.871&0.934&0.949&0.962&0.968&0.975&0.982&0.978&0.974&0.978&0.333\\\hline
\textbf{4}&0.833&0.905&0.931&0.949&0.948&0.964&0.963&0.968&0.969&0.977&0.250\\\hline
\textbf{5}&0.800&0.887&0.923&0.950&0.945&0.951&0.953&0.961&0.961&0.969&0.200\\\hline
\textbf{6}&0.760&0.880&0.929&0.932&0.937&0.941&0.948&0.952&0.950&0.973&0.167\\\hline
\textbf{7}&0.755&0.851&0.909&0.924&0.937&0.943&0.937&0.957&0.960&0.957&0.143\\\hline
\textbf{8}&0.722&0.841&0.898&0.911&0.932&0.941&0.938&0.947&0.952&0.955&0.125\\\hline
\textbf{9}&0.683&0.855&0.882&0.905&0.915&0.931&0.932&0.944&0.952&0.955&0.111\\\hline
\textbf{10}&0.701&0.827&0.882&0.910&0.923&0.923&0.934&0.935&0.943&0.935&0.100\\\hline
\end{tabular}
\caption{Profile length vs. accuracy for different number of authors (text length = 25,000)}
\label{table_pro_v_acc_25}
\end{table*}

\begin{table*}
\centering
\begin{tabular}{|c|l|l|l|l|l|l|l|l|l|l|l|l|l|l|l|l|l|}
\hline
\multirow{2}{*}{Nr. of authors} &\multicolumn{10}{|c|}{Number of words in profile (thousands)} & \multirow{2}{*}{Rand.}\\
\cline{2-11}
&\textbf{10}&\textbf{20}&\textbf{30}&\textbf{40}&\textbf{50}&\textbf{60}&\textbf{70}&\textbf{80}&\textbf{90}&\textbf{100}&\\\hline
\textbf{2}&0.863&0.930&0.932&0.945&0.928&0.952&0.942&0.907&0.942&0.967&0.500\\\hline
\textbf{3}&0.821&0.884&0.886&0.890&0.910&0.901&0.943&0.912&0.911&0.914&0.333\\\hline
\textbf{4}&0.728&0.833&0.849&0.862&0.892&0.867&0.888&0.905&0.882&0.885&0.250\\\hline
\textbf{5}&0.698&0.819&0.825&0.839&0.862&0.884&0.859&0.865&0.882&0.893&0.200\\\hline
\textbf{6}&0.673&0.754&0.789&0.798&0.832&0.837&0.863&0.870&0.896&0.878&0.167\\\hline
\textbf{7}&0.616&0.754&0.806&0.838&0.812&0.848&0.859&0.832&0.873&0.868&0.143\\\hline
\textbf{8}&0.600&0.720&0.748&0.820&0.805&0.831&0.831&0.854&0.857&0.850&0.125\\\hline
\textbf{9}&0.587&0.718&0.767&0.781&0.796&0.809&0.833&0.850&0.843&0.847&0.111\\\hline
\textbf{10}&0.556&0.693&0.737&0.753&0.805&0.827&0.829&0.824&0.843&0.846&0.100\\\hline
\end{tabular}
\caption{Profile length vs. accuracy for different number of authors (text length = 5,000)}
\label{table_pro_v_acc_5}
\end{table*}

\begin{table*}
\centering
\begin{tabular}{|c|l|l|l|l|l|l|l|l|l|l|l|l|l|l|l|l|l|}
\hline
\multirow{2}{*}{Nr. of authors} &\multicolumn{10}{|c|}{Number of words in profile (thousands)} & \multirow{2}{*}{Rand.} \\
\cline{2-11}
&\textbf{10}&\textbf{20}&\textbf{30}&\textbf{40}&\textbf{50}&\textbf{60}&\textbf{70}&\textbf{80}&\textbf{90}&\textbf{100}&\\\hline
\textbf{2}&0.738&0.788&0.747&0.823&0.803&0.803&0.802&0.800&0.812&0.793&0.500\\\hline
\textbf{3}&0.599&0.698&0.690&0.737&0.713&0.744&0.724&0.726&0.757&0.698&0.333\\\hline
\textbf{4}&0.528&0.638&0.640&0.672&0.658&0.663&0.656&0.663&0.651&0.707&0.250\\\hline
\textbf{5}&0.491&0.561&0.598&0.627&0.686&0.621&0.633&0.661&0.674&0.632&0.200\\\hline
\textbf{6}&0.469&0.549&0.578&0.593&0.626&0.594&0.598&0.617&0.606&0.582&0.167\\\hline
\textbf{7}&0.420&0.469&0.539&0.551&0.583&0.564&0.603&0.593&0.583&0.598&0.143\\\hline
\textbf{8}&0.392&0.454&0.544&0.540&0.572&0.551&0.583&0.589&0.563&0.599&0.125\\\hline
\textbf{9}&0.385&0.449&0.489&0.528&0.519&0.556&0.551&0.580&0.560&0.576&0.111\\\hline
\textbf{10}&0.353&0.410&0.466&0.480&0.506&0.536&0.529&0.542&0.556&0.553&0.100\\\hline
\end{tabular}
\caption{Profile length vs. accuracy for different number of authors (text length = 1,000)}
\label{table_pro_v_acc_1}
\end{table*}

The profile length is defined as the total number of words, function or otherwise, used to construct the profile. To study the effect of varying profile lengths we fix $\alpha = 0.75$, $D = 10$, and vary the length of author profiles from 10,000 to 100,000 words in increments of 10,000 words. For each profile length, we attribute texts containing 25,000, 5,000 and 1,000 words, and for each given combination of profile and text length, we consider problems ranging from binary attribution to attribution between ten authors. To build profiles, we use ten texts of the same length randomly chosen among all the texts written by a given author. The length of each excerpt is such that the ten pieces add up to the desired profile length. E.g., to build a profile of length 50,000 words for Melville, we randomly pick ten excerpts of 5,000 words each among all the texts written by him. For the texts to be attributed, however, we always select contiguous extracts of the desired length. E.g., for texts of length 25,000 words, we randomly pick excerpts of this length written by some author -- as opposed to the selection of ten pieces of different origin we do for the profiles. This resembles the usual situation where the profiles are built from several sources but the texts to attribute correspond to a single literary creation. For a given profile size and number of authors, several attribution experiments were ran by randomly choosing the set of authors among those from the 19th century \cite{Corpus_texts_2} and randomly choosing the texts forming the profiles. The amount of attribution experiments was chosen large enough to ensure that every accuracy value in tables \ref{table_pro_v_acc_25} - \ref{table_pro_v_acc_1} is based on the attribution of at least 600 texts. 

The accuracy results of attributing a text of 25,000 words are stated in Table \ref{table_pro_v_acc_25}. This word length is equivalent to around 60 pages of a midsize paperback novel -- i.e., a novella, or a few book chapters -- or the typical length of a Shakespeare play. In the last column of the table we inform the expected accuracy of random attribution between the candidate authors. The difference between the accuracies of the last column and the rest of the table indicates that WANs \emph{do} carry stylometric information useful for authorship attribution. Overall, attribution of texts with 25,000 words can be done with high accuracy even when attributing among a large number of authors if reasonably large corpora are available to build author profiles with 60,000 to 100,000 words. E.g., for a profile containing 40,000 words, our method achieves an accuracy of 0.985 for binary attributions whereas the corresponding random accuracy is 0.5. As expected, accuracy decreases when the number of candidate authors increases. E.g., for profiles of 80,000 words, an accuracy of 0.986 is obtained for binary attributions whereas an accuracy of 0.935 is obtained when the pool of candidates contains ten authors. Observe that accuracy does not monotonically decrease when increasing the candidate authors due to the noise introduced by the random selection of authors and texts. 

Accuracy increases with longer profiles. E.g., when performing attributions of 25,000 word texts among 6 authors, the accuracy obtained for profiles of length 10,000 is 0.760 whereas the accuracy obtained for profiles of length 60,000 is 0.941. There is a saturation effect concerning the length of the profile that depends on the number of authors being considered. For binary attributions there is no major increase in accuracy beyond profiles of length 30,000. However, when the number of candidate authors is 7, accuracy stabilizes for profiles of length in the order of 80,000 words. There seems to be little benefit in using profiles containing more than 100,000 words, which corresponds to a short novel of about 250 pages.

Correct attribution rates of shorter excerpts containing 5,000 words are shown in Table \ref{table_pro_v_acc_5} for the same profile lengths and number of candidate authors considered in Table \ref{table_pro_v_acc_25}. A text of this length corresponds to about 13 pages of a novel -- something in the order of the chapter of a book -- or an act in a Shakespeare play. When considering these shorter texts, acceptable classification accuracy is achieved except for very short profiles and large number of authors, while reliable attribution requires a small number of candidate authors or a large profile. E.g., attribution between three authors with profiles of 70,000 words has an average accuracy of 0.943. While smaller than the corresponding correct attribution rate of 0.982 for texts of length 25,000 words, this is still a respectable number. To achieve an accuracy in excess of 0.9 for the case of three authors we need a profile of at least 50,000 words.

For very short texts of 1,000 words, which is about the length of an opinion piece in a newspaper, a couple pages in a novel, or a scene in a Shakespeare play, we can provide indications of authorship but cannot make definitive claims. As shown in Table \ref{table_pro_v_acc_1}, the best accuracies are for binary attributions that hover at around 0.8 when we use profiles longer than 40,000 words. For attributions between more than 2 authors, maximum correct attribution rates are achieved for profiles containing 90,000 or 100,000 words and range from 0.757 for the case of three authors to 0.556 when considering ten authors. These rates are markedly better than random attribution but not sufficient for definitive statements. The results can be of use as circumstantial evidence in support of attribution claims substantiated by further proof.

\begin{table*}
\centering
\begin{tabular}{|c|l|l|l|l|l|l|l|l|l|l|l|l|l|}
\hline
\multirow{2}{*}{Nr. of authors} &\multicolumn{12}{|c|}{Number of words in texts (thousands)} & \multirow{2}{*}{Rand.} \\
\cline{2-13}
&\textbf{1}&\textbf{2}&\textbf{3}&\textbf{4}&\textbf{5}&\textbf{6}&\textbf{8}&\textbf{10}&\textbf{15}&\textbf{20}&\textbf{25}&\textbf{30}&\\\hline
\textbf{2}&0.840&0.917&0.925&0.938&0.940&0.967&0.958&0.977&0.967&0.989&0.988&0.986&0.500\\\hline
\textbf{3}&0.789&0.873&0.890&0.919&0.913&0.932&0.936&0.956&0.952&0.979&0.979&0.975&0.333\\\hline
\textbf{4}&0.736&0.842&0.870&0.902&0.906&0.933&0.937&0.952&0.965&0.970&0.973&0.974&0.250\\\hline
\textbf{5}&0.711&0.797&0.858&0.874&0.891&0.906&0.924&0.925&0.955&0.971&0.980&0.964&0.200\\\hline
\textbf{6}&0.690&0.796&0.828&0.886&0.884&0.911&0.919&0.922&0.944&0.957&0.969&0.961&0.167\\\hline
\textbf{7}&0.633&0.730&0.814&0.855&0.874&0.890&0.910&0.911&0.928&0.947&0.956&0.951&0.143\\\hline
\textbf{8}&0.602&0.740&0.811&0.846&0.882&0.887&0.915&0.910&0.930&0.944&0.957&0.963&0.125\\\hline
\textbf{9}&0.607&0.721&0.774&0.826&0.845&0.870&0.889&0.890&0.918&0.948&0.951&0.953&0.111\\\hline
\textbf{10}&0.578&0.731&0.792&0.816&0.842&0.855&0.872&0.893&0.921&0.933&0.942&0.961&0.100\\\hline
\end{tabular}
\caption{Text length vs. accuracy for different number of authors (profile length = 100,000)}
\label{table_text_v_acc_100}
\end{table*}

\begin{table*}
\centering
\begin{tabular}{|c|l|l|l|l|l|l|l|l|l|l|l|l|l|}
\hline
\multirow{2}{*}{Nr. of authors} &\multicolumn{12}{|c|}{Number of words in texts (thousands)} & \multirow{2}{*}{Rand.} \\
\cline{2-13}
&\textbf{1}&\textbf{2}&\textbf{3}&\textbf{4}&\textbf{5}&\textbf{6}&\textbf{8}&\textbf{10}&\textbf{15}&\textbf{20}&\textbf{25}&\textbf{30}&\\\hline
\textbf{2}&0.812&0.850&0.903&0.912&0.913&0.912&0.938&0.945&0.918&0.964&0.964&0.969&0.500\\\hline
\textbf{3}&0.760&0.797&0.858&0.899&0.887&0.918&0.920&0.918&0.919&0.938&0.929&0.928&0.333\\\hline
\textbf{4}&0.670&0.747&0.813&0.852&0.868&0.887&0.889&0.906&0.918&0.915&0.900&0.913&0.250\\\hline
\textbf{5}&0.621&0.721&0.749&0.813&0.823&0.819&0.859&0.878&0.876&0.887&0.889&0.893&0.200\\\hline
\textbf{6}&0.557&0.681&0.754&0.782&0.799&0.831&0.852&0.866&0.871&0.879&0.881&0.872&0.167\\\hline
\textbf{7}&0.493&0.610&0.674&0.706&0.731&0.770&0.798&0.807&0.828&0.862&0.867&0.858&0.143\\\hline
\textbf{8}&0.467&0.623&0.675&0.721&0.741&0.769&0.790&0.826&0.822&0.857&0.841&0.857&0.125\\\hline
\textbf{9}&0.474&0.574&0.656&0.672&0.710&0.734&0.781&0.783&0.813&0.845&0.837&0.841&0.111\\\hline
\textbf{10}&0.433&0.535&0.612&0.663&0.684&0.706&0.752&0.772&0.836&0.840&0.851&0.848&0.100\\\hline
\end{tabular}
\caption{Text length vs. accuracy for different number of authors (profile length = 20,000)}
\label{table_text_v_acc_20}
\end{table*}

\begin{table*}
\centering
\begin{tabular}{|c|l|l|l|l|l|l|l|l|l|l|l|l|l|}
\hline
\multirow{2}{*}{Nr. of authors} &\multicolumn{12}{|c|}{Number of words in texts (thousands)} & \multirow{2}{*}{Rand.} \\
\cline{2-13}
&\textbf{1}&\textbf{2}&\textbf{3}&\textbf{4}&\textbf{5}&\textbf{6}&\textbf{8}&\textbf{10}&\textbf{15}&\textbf{20}&\textbf{25}&\textbf{30}&\\\hline
\textbf{2}&0.672&0.740&0.747&0.707&0.803&0.823&0.788&0.848&0.820&0.802&0.827&0.832&0.500\\\hline
\textbf{3}&0.547&0.623&0.626&0.653&0.744&0.669&0.712&0.757&0.736&0.764&0.734&0.733&0.333\\\hline
\textbf{4}&0.452&0.487&0.528&0.597&0.652&0.623&0.623&0.662&0.682&0.661&0.632&0.694&0.250\\\hline
\textbf{5}&0.403&0.510&0.535&0.538&0.505&0.573&0.618&0.592&0.681&0.606&0.638&0.570&0.200\\\hline
\textbf{6}&0.372&0.457&0.480&0.485&0.529&0.518&0.545&0.577&0.605&0.631&0.599&0.601&0.167\\\hline
\textbf{7}&0.349&0.382&0.460&0.469&0.475&0.504&0.522&0.539&0.528&0.568&0.588&0.562&0.143\\\hline
\textbf{8}&0.302&0.390&0.453&0.440&0.473&0.510&0.465&0.517&0.541&0.530&0.534&0.549&0.125\\\hline
\textbf{9}&0.296&0.347&0.370&0.427&0.477&0.439&0.485&0.492&0.506&0.530&0.557&0.532&0.111\\\hline
\textbf{10}&0.254&0.337&0.373&0.405&0.413&0.427&0.455&0.487&0.480&0.460&0.443&0.463&0.100\\\hline
\end{tabular}
\caption{Text length vs. accuracy for different number of authors (profile length = 5,000)}
\label{table_text_v_acc_5}
\end{table*}

\subsection{Varying Text Length}\label{sec:varying_text_length}

In this section we analyze the effect of text length in attribution accuracy for varying profile lengths and number of candidate authors. Using $\alpha = 0.75$ and $D = 10$, we consider profiles of length 100,000, 20,000 and 5,000 words and vary the number of candidate authors from two to ten. The text lengths considered are 1,000 words to 6,000 words in 1,000 word increments, 8,000 words, and 10,000 to 30,000 words in 5,000 word increments. The fine resolution for short texts permits estimating the shortest texts that can be attributed accurately. As in Section \ref{sec:varying_profile_length}, for every combination of number of authors and text length, enough independent attribution experiments were performed to ensure that every accuracy value in tables \ref{table_text_v_acc_100} - \ref{table_text_v_acc_5} is based on at least 600 attributions.

For profiles of length 100,000 words, the results are reported in Table \ref{table_text_v_acc_100}. As done in tables \ref{table_pro_v_acc_25}-\ref{table_pro_v_acc_1}, we state the expected accuracy of random attribution in the last column of the table. Accuracies reported towards the right end of the table, i.e. 20,000-30,000 words, correspond to the attribution of a dramatic play or around 60 pages of a novel, which we will refer to as long texts. Accuracies for columns in the middle of the table, i.e. 5,000-8,000 words, correspond to an act in a dramatic play or between 12 and 20 pages of a novel, which we will refer to as medium texts. The left columns of this table, i.e. 1,000-3,000 words, correspond to a scene in a play, 2 to 7 pages in a novel or an article in a newspaper, which we will refer to as short texts. For the attribution of long texts, we achieve a mean accuracy of 0.988 for binary attributions which decreases to an average accuracy of 0.945 when the number of candidate authors is increased to ten. For medium texts, the decrease in accuracy is not very significant for binary attributions, with a mean accuracy of 0.955, but the accuracy is reduced to 0.856 for attributions among ten authors. The accuracy is decreased further when attributing short texts, with a mean accuracy of 0.894 for binary attributions and 0.700 for the case with ten candidates. This indicates that when profiles of around 100,000 are available, WANS achieve accuracies over 0.95 for medium to long texts. For short texts, acceptable classification rates are achieved if the number of candidate authors is between two and four.

If we reduce the length of the profiles to 20,000 words, reasonable accuracies are attained for small pools of candidate authors; see Table \ref{table_text_v_acc_20}. E.g, for binary attributions, the range of correct classification varies between 0.812 for texts of 1,000 words to 0.969 for texts with 30,000 words. The first of these numbers means that we can correctly attribute a newspaper opinion piece with accuracy 0.812 if we are given corpora of 20 opinion pieces by the candidate authors. The second of these numbers means that we can correctly attribute a play between two authors with accuracy 0.969 if we are given corpora of 20,000 words by the candidate authors. Further reducing the profile length to 5,000 words results in classification accuracies that are acceptable only when we consider binary attributions and texts of at least 10,000 words; see Table \ref{table_text_v_acc_5}. For shorter texts or larger number of candidate authors, WANs can provide supporting evidence but not definitive proof.

\begin{figure}
        \centering
          \includegraphics[width=0.5\textwidth]{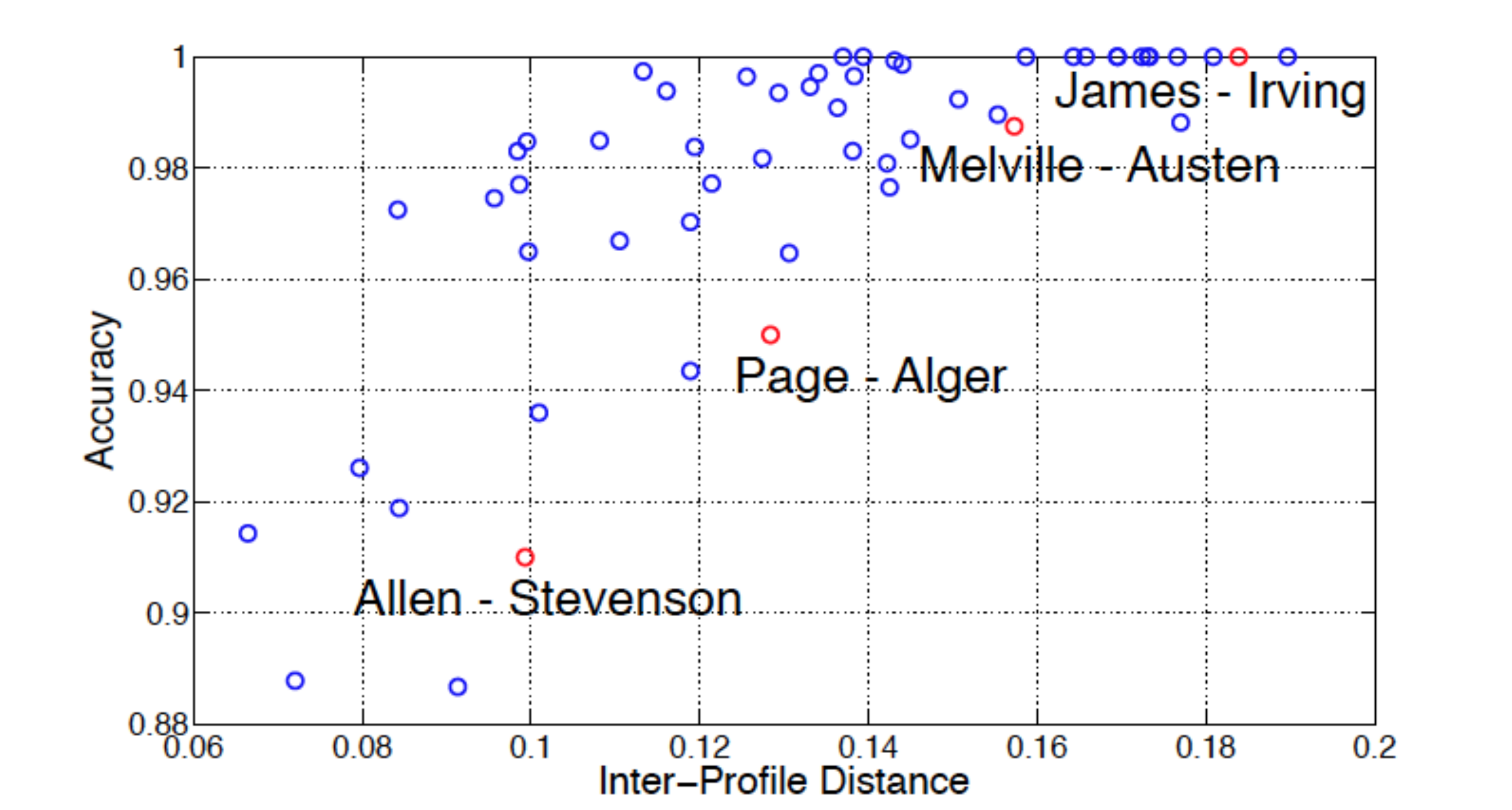}
          \caption{Binary attribution accuracy as a function of the inter-profile dissimilarity. Higher accuracy is attained for attribution between authors which are more dissimilar.}
          \label{fig:accuracy_vs_inter_profile_dist}
\end{figure}


\subsection{Inter-Profile Dissimilarities}\label{sec:inter_profile_distance}

Besides the number of candidate authors and the length of the texts and profiles, the correct attribution of a text is also dependent on the similarity of the writing styles of the authors themselves. Indeed, repeated binary attributions between Henry James and Washington Irving with random generation of 100,000 word profiles yield a perfect accuracy of 1.0 on the classification of 400 texts of 10,000 words each. The same exercise when attributing between Grant Allen and Robert Louis Stevenson yields a classification rate of 0.91. This occurs because the stylometric fingerprints of Allen and Stevenson are harder to distinguish than those of James and Irving.

Dissimilarity of writing styles can be quantified by computing the relative entropies between the profiles [cf. \eqref{eqn_def_H_in_practice}]. Since relative entropies are asymmetric, i.e., $H(P_1, P_2) \neq H(P_2, P_1)$ in general, we consider the average of the two relative entropies between two profiles as a measure of their dissimilarity. For each pair of authors, the relative entropy is computed based on the set of function words chosen adaptively to maximize the cross validation accuracy. For the 100,000 word profiles of James and Irving, the inter-profile dissimilarity resulting from the average of relative entropies is 0.184. The inter-profile dissimilarity between Allen and Stevenson is 0.099. This provides a formal measure of similarity of writing styles which explains the higher accuracy of attributions between James and Irving with respect to attributions between Allen and Stevenson.
 
The correlation between inter-profile dissimilarities and attribution accuracy is corroborated by Fig. \ref{fig:accuracy_vs_inter_profile_dist}. Each point in this plot corresponds to the selection of two authors at random from our pool of 21 authors from the 19th century. For each pair we select ten texts of 10,000 words each to generate profiles of length 100,000 words. We then attribute ten of the remaining excerpts of length 10,000 words of each of these two authors among the two profiles and record the correct attribution rate as well as the dissimilarity between the random profiles generated. The process is repeated twenty times for these two authors to produce the average dissimilarity and accuracy that yield the corresponding point in Fig. \ref{fig:accuracy_vs_inter_profile_dist}. E.g., consider two randomly chosen authors for which we have 50 excerpts of 10,000 word available. We select ten random texts to form a profile and attribute 20 out of the remaining 80 excerpts -- 10 for each author. After repeating this procedure twenty times we get the average accuracy of attributing 400 texts of length 10,000 words between the two authors. 


Besides the positive correlation between inter-profile dissimilarities and attribution accuracies, Fig. \ref{fig:accuracy_vs_inter_profile_dist} shows that classification is perfect for 11 out of 12 instances where the inter-profile dissimilarity exceeds 0.16. Errors are rare for profile dissimilarities between 0.10 and 0.16 since correct classifications average 0.984 and account for at least 0.96 of the attribution results in all but three outliers. For pairs of authors with dissimilarities smaller than 0.1 the average accuracy is 0.942.

\section{Meta attribution studies}

WANs can also be used to study problems other than attribution between authors. In this section we demonstrate that WANs carry information about time periods, the genre of the composition, and the gender of the authors. We also illustrate the use of WANs in detecting collaborations.

\begin{figure*}
        \centering
        \begin{subfigure}[b]{0.5\textwidth}
          \hspace{-0.2in} \includegraphics[width=\textwidth]{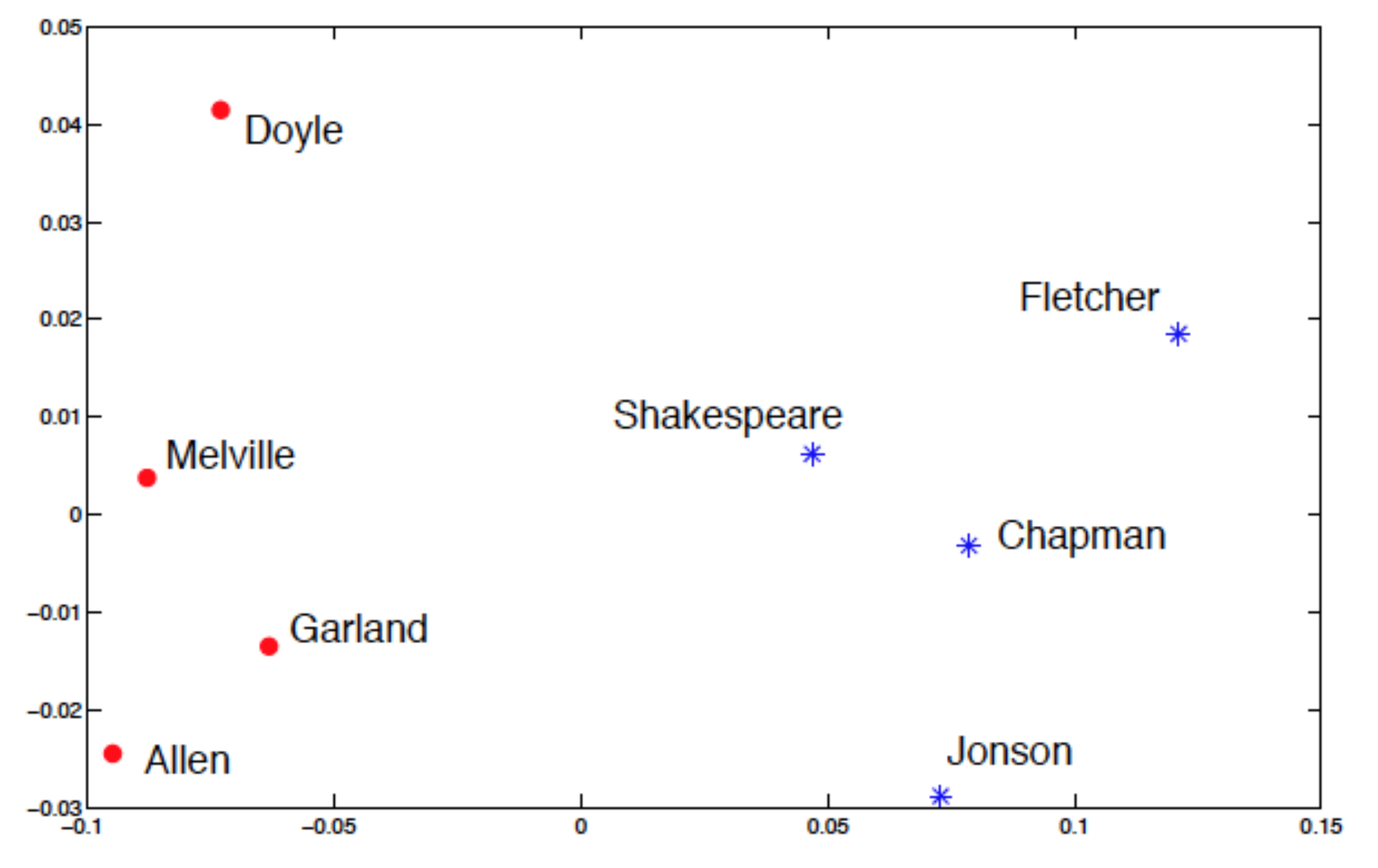}
          \caption{MDS plot for authors of different time periods.}
          \label{fig:time_mds}
        \end{subfigure}%
        \begin{subfigure}[b]{0.43\textwidth}
          \includegraphics[width=\textwidth]{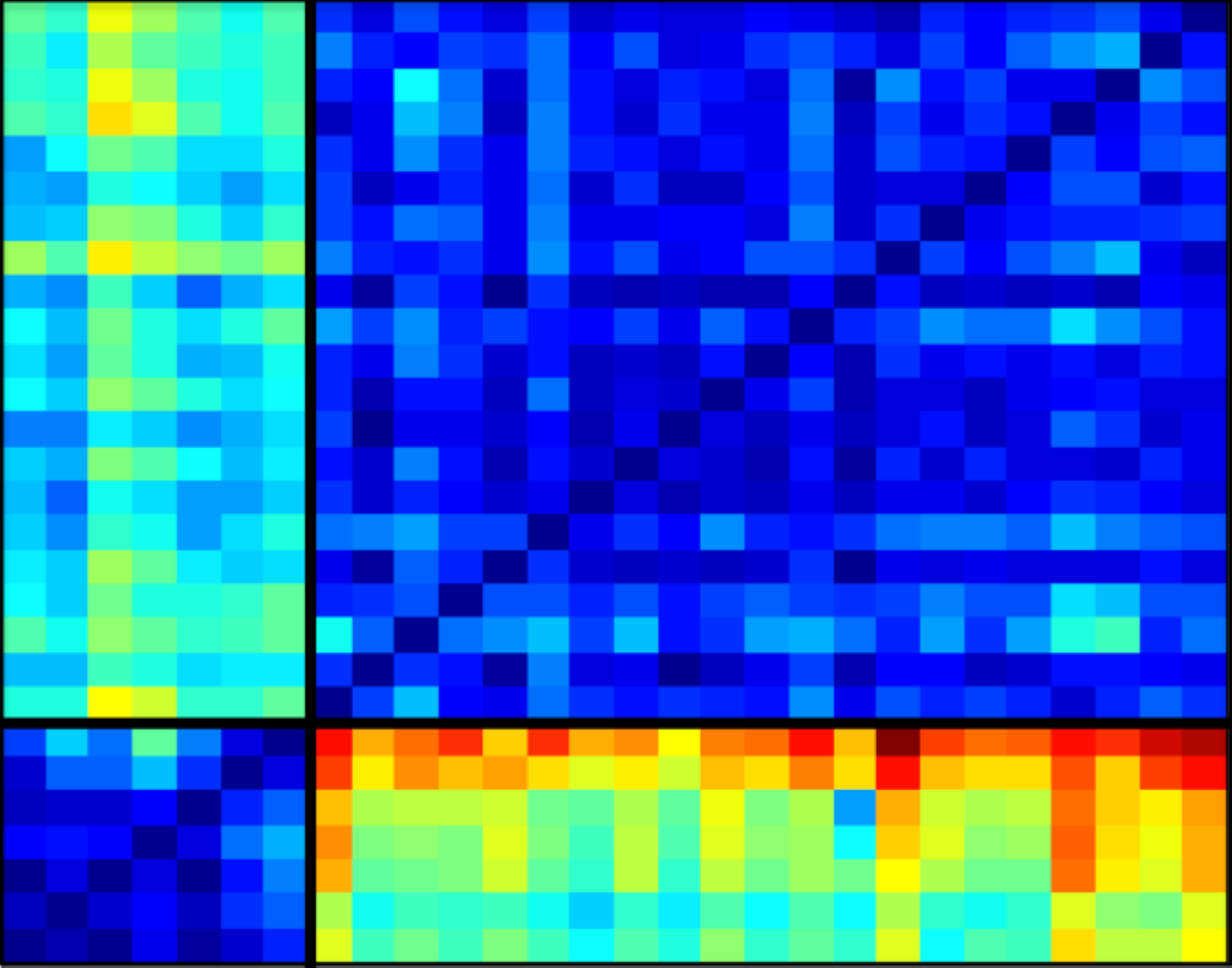}
          \caption{Heat map of inter-profile relative entropies. }
          \label{fig:time_heatmap}
        \end{subfigure}
        \caption{(a) Authors from the early 17th century are depicted as blue stars while authors from the 19th century are depicted as red dots. Inter-profile dissimilarities are small within the groups and large between them. (b) High inter-profile relative entropies are illustrated with warmer colors. Two groups of authors with small inter-profile relative entropies are apparent: the first seven correspond to 17th century authors and the rest to 19th century authors.}
        \label{fig:time_comparison}
\end{figure*}

\subsection{Time}\label{sec:time}

WANs carry information about the point in time in which a text was written. If we build random profiles of 200,000 words for Shakespeare, Chapman, and Melville and compute the inter-profile dissimilarity as in Section \ref{sec:inter_profile_distance}, we obtain a dissimilarity of 0.04 between Shakespeare and Chapman and of 0.17 between Shakespeare and Melville. Since inter-profile dissimilarity is a measure of difference in style, this values are reasonable given that Shakespeare and Chapman were contemporaries but Melville lived more than two centuries after them.

To further illustrate this point, in Fig. \ref{fig:time_mds} we plot a two dimensional MDS representation of the dissimilarity between eight authors whose profiles were built with all their available texts in our corpus \cite{Corpus_texts_2}. Four of the profiles correspond to early 17th century authors -- Shakespeare, Chapman, Jonson, and Fletcher -- and are represented by blue stars while the other four -- Doyle, Melville, Garland, and Allen -- correspond to 19th century authors and are represented by red dots. Notice that authors tend to have a smaller distance with their contemporaries and a larger distance with authors from other periods. This fact is also illustrated by the heat map of inter-profile relative entropies in Fig. \ref{fig:time_heatmap} where bluish colors represent smaller entropies. Since heat maps allow the representation of asymmetric data, we directly plot the relative entropies instead of the symmetrized inter-profile dissimilarities. The first 7 rows and columns correspond to authors of the 17th century whereas the remaining 21 correspond to authors of the 19th century, where profiles were built with all the available texts. Notice that the blocks of blue color along the diagonal are in perfect correspondence with the time period of the authors, verifying that WANs can be used to determine the time in which a text was written. The average entropies among authors of the 17th century and among those of the 19th century are 0.096 and 0.098 respectively, whereas the average entropies between authors of different epochs is 0.273. I.e., the relative entropy between authors of different epochs almost triples that of authors belonging to the same time period.

\begin{table}[t]
\centering
{\normalsize
 \begin{tabular}{|c|c|c|}
\cline{2-3} \multicolumn{1}{c|}{} & Marlowe & Chapman \\ 
\hline
Shakespeare (Com.) & 11.6 & 7.7 \\
\hline 
Shakespeare (His.) & 7.6 & 9.3 \\
 \hline
 \end{tabular}
 }
 \caption{Inter-profile dissimilarities (x100) between authors of different genres.}
\label{tab:genre_distance}
\end{table}

\subsection{Genre}\label{sec:genre}

\begin{figure}
        \centering
          \includegraphics[width=0.43\textwidth]{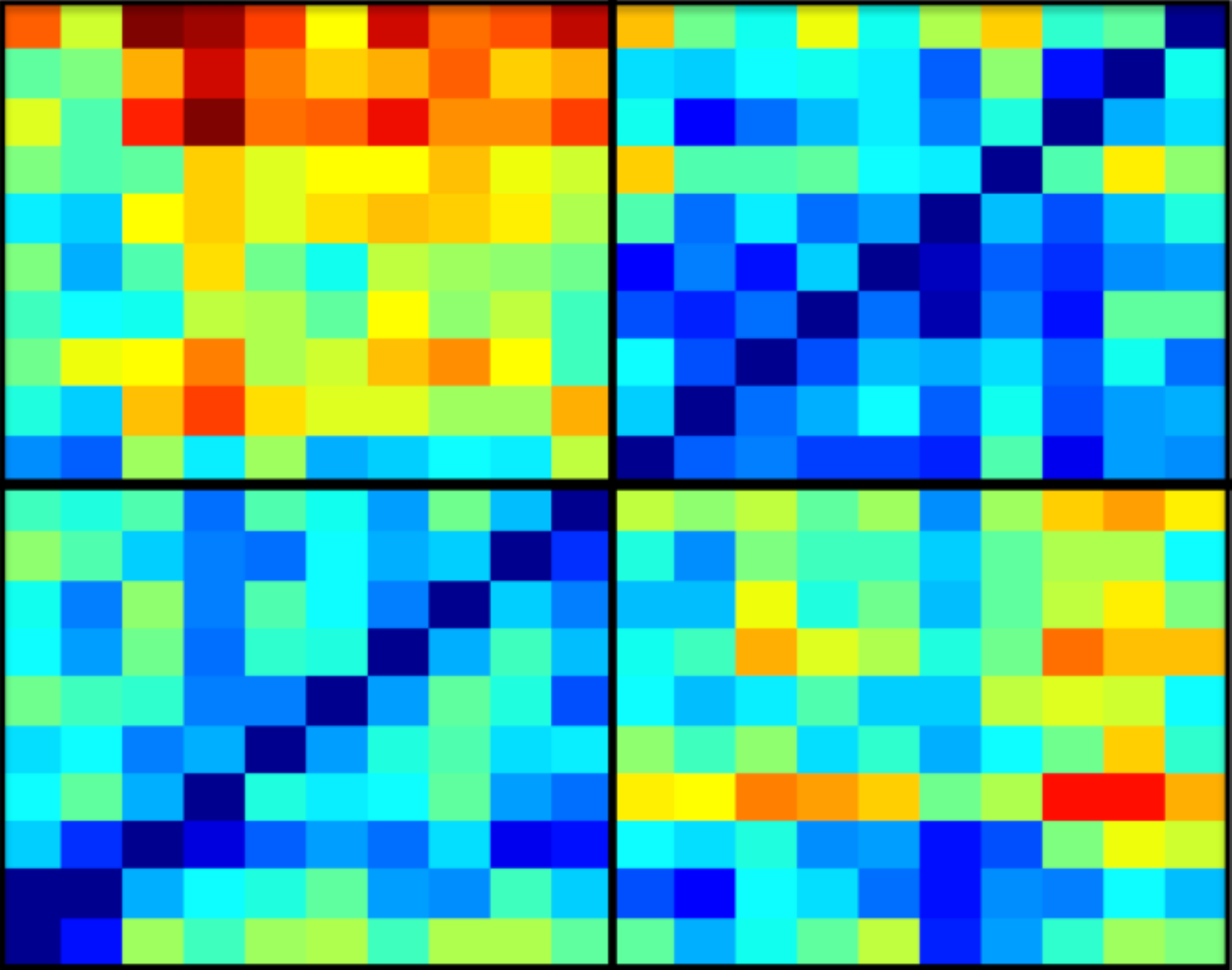}
          \caption{Heat map of relative entropies between 20 Shakespeare extracts. The first 10 texts correspond to history plays while the last 10 correspond to comedy plays. Relative entropies within texts of the same genre are smaller than across genres.}
          \label{fig:genre_heatmap}
\end{figure}

Even though function words by themselves do not carry content, WANs constructed from a text contain, rather surprisingly, information about its genre. We illustrate this fact in Fig. \ref{fig:genre_heatmap}, where we present the relative entropy between 20 pieces of texts written by Shakespeare of length 20,000 words, where 10 of them are history plays -- e.g., \emph{Richard II, King John, Henry VIII} -- and 10 of them are comedies -- e.g., \emph{The Tempest, Measure for Measure, The Merchant of Venice}. As in Fig. \ref{fig:time_heatmap}, bluish colors in Fig. \ref{fig:genre_heatmap} represent smaller relative entropies. Two blocks along the diagonal can be distinguish that coincide with the plays of the two different genres. Indeed, if we sequentially extract one text from the group and attribute it to a genre by computing the average relative entropies to the remaining histories and comedies, the 20 pieces are correctly attributed to their genre.

More generally, inter-profile dissimilarities between authors that write in the same genre tend to be smaller than between authors that write in different genres. As an example, in Table \ref{tab:genre_distance} we compute the dissimilarity between two Shakespeare profiles -- one built with comedies and the other with histories -- and two contemporary authors: Marlowe and Chapman. All profiles contain 100,000 words formed by randomly picking 10 extracts of 10,000 words. Marlowe never wrote a comedy and mainly focused on histories -- \emph{Edward II, The Massacre at Paris} -- and tragedies -- \emph{The Jew of Malta, Dido} --, while the majority of Chapman's plays are comedies -- \emph{All Fools, May Day}. Genre choice impacts the inter-profile dissimilarity since the comedy profile of Shakespeare is closer to Chapman than to Marlowe and vice versa for the history profile of Shakespeare. The inter-profile dissimilarity between Shakespeare profiles is 6.2, which is still smaller than any dissimilarity in Table \ref{tab:genre_distance}. This points towards the conclusion that the identity of the author is the main determinant of the writing style but that the genre of the text being written also contributes to the word choice. In general, two texts of the same author but different genres are more similar than two texts of the same genre but different authors which, in turn, are more similar than two texts of different authors and genres.

\begin{table}[t]
\centering
{\normalsize
 \begin{tabular}{|c|c|c|c|c|c|c|}
 \hline
Sh. & Jon. & Fle. & Mid. & Cha. & Marl. \\ 
\hline
{\bf 19.1} & 20.0 & {\bf 18.2} & 20.2 & 19.5 & 20.9 \\
 \hline
 \end{tabular}
 }
 \caption{Relative entropies from \emph{Two Noble Kinsmen} to different profiles (x100).}
\label{tab:tnk_distance_pure}
\end{table}

\begin{table}[t]
\centering
{\normalsize
 \begin{tabular}{|c|c|c|c|c|c|c|}
 \cline{2-7}
\multicolumn{1}{c|}{} & Sh. & Jon. & Fle. & Mid. & Cha. & Marl. \\ \hline
Sh.     & 19.1  &  19.2    & {\bf 17.9}  &  19.0    & 19.1  &  19.3 \\ \hline
Jon.     & 19.2   & 20.0    & 18.4   & 19.5    & 19.3   & 19.3 \\ \hline
Fle.     & {\bf 17.9}   & 18.4    & 18.2    & 18.4    & 18.2    & 18.1 \\ \hline
Mid.     & 19.0    & 19.5    & 18.4     & 20.2    & 19.4    & 18.9  \\ \hline
Cha.     & 19.1    & 19.3     & 18.2    & 19.4    & 19.5    & 19.4  \\ \hline
Marl.     & 19.3    & 19.3   & 18.1    & 18.9   & 19.4    & 20.9 \\ \hline

 \end{tabular}
 }
 \caption{Relative entropies from \emph{Two Noble Kinsmen} to hybrid profiles composed of two authors (x100).}
\label{tab:tnk_distance_hybrid}
\end{table}

\begin{table}[t]
\centering
{\normalsize
 \begin{tabular}{|c|c|c|c|c|c|c|}
 \cline{2-7}
\multicolumn{1}{c|}{} & Sh. & Jon. & Fle. & Mid. & Cha. & Marl. \\ \hline
Sh.     & 17.6  & 16.8   &17.3 &  16.7   &17.1 &  18.2    \\ \hline
Jon.     & 16.8  & {\bf 16.8}   &17.0  & {\bf 16.5}   & {\bf 16.7}  & 17.3  \\ \hline
Fle.     & 17.3  & 17.0   &18.7   &17.6   &17.4   &17.9   \\ \hline
Mid.     & 16.7  & {\bf 16.5}  & 17.6   &17.6   &16.9   &17.1   \\ \hline
Cha.     & 17.1  & {\bf 16.7}   &17.4   &16.9   & {\bf 17.5}   &17.8  \\ \hline
Marl.     & 17.4  & 17.1 &  17.6   &17.3&   17.4   & 18.1 \\ \hline   
 \end{tabular}
 }
 \caption{Relative entropies from \emph{Eastward Ho} to hybrid profiles composed of two authors (x100).}
\label{tab:eastward_distance_hybrid}
\end{table}


\subsection{Gender}\label{sec:gender}

Word usage can be used for author profiling \cite{Argamonetal09} and, in particular, to infer the gender of an author from the written text. To illustrate this, we divide the 21 authors from the 19th century \cite{Corpus_texts_2} into females -- five of them -- and males. We pick a gender at random and pick an excerpt of 10,000 words from any author of the selected gender. We then build two 100,000 words profiles, one containing pieces of texts written by male authors and the other by female authors. In order to avoid bias, we do not include any text of the author from which the text to attribute was chosen in the gender profiles. We then attribute the chosen text between the two gender profiles. After repeating this procedure 5,000 times, we obtain a mean accuracy of 0.63. Although this accuracy is lower than state-of-the-art gender profiling methods \cite{Koppeletal02}, the difference with random attribution, i.e. accuracy of 0.5, validates the fact that WANs carry gender information about the authors.

\begin{table*}
\centering
{\normalsize
 \begin{tabular}{|x{2cm}|x{1.5cm}|x{1.5cm}|x{1.5cm}|x{1.5cm}|x{1.5cm}|x{1.5cm}|x{1.5cm}|x{1.5cm}|}
 \hline
Nr. of authors & N. Bayes & 1-NN & 3-NN & DT-gdi & DT-ce & SVM & WAN & Voting \tabularnewline \hline
2     &  2.6  &  3.5 & 5.2  & 12.2 & 12.2 & 2.7 & {\bf1.6} & 0.9    \tabularnewline \hline
4     &   6.0 & 9.2 & 12.4 & 25.3 & 25.5 & 6.8 & {\bf4.6} &  3.3 \tabularnewline \hline
6     &   8.1 & 11.7 & 15.2 & 31.9 & 32.2 & 7.9 & {\bf5.3} & 3.8   \tabularnewline \hline
8     &  9.6 & 15.4 & 19.2 & 36.4 & 37.2 & 11.1 & {\bf6.7} & 5.2   \tabularnewline \hline
10     &   10.8 & 16.7 & 21.4 & 42.1 & 42.1 & 11.5 & {\bf8.3} & 6.0 \tabularnewline \hline   
 \end{tabular}
 }
 \caption{Error rates in \% achieved by different methods for profiles of 100,000 words and texts of 10,000 words. The WANs  achieve the smallest error rate among the methods considered separately. Voting decreases the error even further by combining the relational data of the WANs with the frequency data of other methods. }
\label{tab:compare_methods_1}
\end{table*}

\subsection{Collaborations}\label{sec:detecting_collaborations}

WANs can also be used for the attribution of texts written collaboratively between two or more authors. Since collaboration was a common practice for playwrights in the early 17th century, we consider the attribution of Early Modern English plays \cite{Corpus_texts_2}. For a given play, we compute its relative entropy to six contemporary authors -- Shakespeare, Jonson, Fletcher, Middleton, Chapman, and Marlowe -- by generating 50 random profiles for each author of length 80,000 words and averaging the 50 entropies to obtain one representative number. We do not consider Peele in the analysis due to the short total length of available texts. 

When two authors collaborate to write a play, the resulting word adjacency network is close to the profiles of both authors, even though these profiles are built with plays of their sole authorship. As an example, consider the play \emph{Two Noble Kinsmen} which is an accepted collaboration between Fletcher and Shakespeare \cite{DEEP}. In Table \ref{tab:tnk_distance_pure}, we present the relative entropies between the play and the six analyzed authors. Notice that the two minimum entropies correspond to those who collaborated in writing it. 

Collaboration can be further confirmed by the construction of hybrid profiles, i.e. profiles built containing 40,000 words of two different authors. Each entry in Table \ref{tab:tnk_distance_hybrid} corresponds to the relative entropy from \emph{Two Noble Kinsmen} to a hybrid profile composed by the authors in the row and column of that entry. 
Notice that the diagonal of Table \ref{tab:tnk_distance_hybrid} corresponds to profiles of sole authors and, thus, coincides with Table \ref{tab:tnk_distance_pure}. The smallest relative entropy in Table \ref{tab:tnk_distance_hybrid} is achieved by the hybrid profile composed by Fletcher and Shakespeare, which is consistent with the accepted attribution of the play.

The attribution between hybrid profiles is not always accurate. For example, consider the play \emph{Eastward Ho} which is a collaboration between three authors, two of which are Chapman and Jonson \cite{DEEP}. If we repeat the above procedure and compute the relative entropies between the play and the different pure profiles, we see that in fact the two smallest entropies are achieved for Jonson and Chapman; see the diagonal in Table \ref{tab:eastward_distance_hybrid}. However, the smallest entropy in the whole table is achieved by the hybrid profile composed by Jonson and Middleton. The hybrid profile of Jonson and Chapman, the real authors, achieves an entropy of 16.7, which is the second smallest among all profiles in Table \ref{tab:eastward_distance_hybrid}.


\section{Comparison and Combination with Frequency Based Methods}\label{sec:comparison_combination_frequency_based}

Machine learning tools have been used to solve attribution problems by relying on the frequency of appearance of function words \cite{ZhaoZobel05}. These methods consider the number of times an author uses different function words but, unlike WANs, do not contemplate the order in which the function words appear. The most common techniques include naive Bayes \cite[Chapter 8]{Bishop06}, nearest neighbors (NN) \cite[Chapter 2]{Bishop06}, decision trees (DT) \cite[Chapter 14]{Bishop06}, and support vector machines (SVM) \cite[Chapter 7]{Bishop06}.

In Table \ref{tab:compare_methods_1} we inform the percentage of errors obtained by different methods when attributing texts of 10,000 words among profiles of 100,000 words for a number of authors ranging from two to ten. For a given number of candidate authors, we randomly pick them from the pool of 19th century authors \cite{Corpus_texts_2} and attribute ten excerpts of each of them using the different methods. We then repeat the random choice of authors 100 times and average the error rate. For each of the methods based on function word frequencies, we pick the set of parameters and preprocessing that minimize the attribution error rate. E.g., for SVM the error is minimized when considering a polynomial kernel of degree 3 and normalizing the frequencies by text length. For the nearest neighbors method we consider two strategies based on one (1-NN) and three (3-NN) nearest neighbors as given by the $l_2$ metric in Euclidean space. Also, for decision trees we consider two types of split criteria: the Gini Diversity Index (DT-gdi) and the cross-entropy (DT-ce) \cite{Breiman84}.

The WANs achieve a lower attribution error than frequency based methods; see Table \ref{tab:compare_methods_1}. For binary attributions, naive Bayes and SVM achieve error rates of 2.6\% and 2.7\% respectively and, thus, outperform nearest neighbors and decision trees. However, WANs outperform the aforementioned methods by obtaining an error rate of 1.6\%. This implies a reduction of 38\% in the error rate. For 6 authors, WANs achieve an error rate of 5.3\% that outperform SVMs achieving 7.9\% entailing a 33\% reduction. This trend is consistent across different number of candidate authors, with WANs achieving an average error reduction of 29\% compared with the best traditional machine learning method.

More important than the fact that WANs tend to outperform methods based on word frequencies, is the fact that they carry different stylometric information. Thus, we can combine both methodologies to further increase attribution accuracy. In the last column of Table \ref{tab:compare_methods_1} we inform the error rate of majority voting between WANs and the two best performing frequency based methods, namely, naive Bayes and SVMs. The error rates are consistently smaller than those achieved by WANs and, hence, by the other frequency based methods as well. E.g., for attributions among four authors, voting achieves an error of 3.3\% compared to an error of 4.6\% of WANs. This corresponds to a 28\% reduction in error. Averaging among attributions for different number of candidate authors, majority voting entails a reduction of 30\% compared with WANs. The combination of WANs and function word frequencies halves the attribution error rate with respect to the current state of the art.


\section{Conclusion}\label{sec:conclusion}
Relational data between function words was used as stylometric information to solve authorship attribution problems. Normalized word adjacency networks (WANs) were used as relational structures. We interpreted these networks as Markov chains in order to facilitate their comparison using relative entropies. The accuracy of WANs was analyzed for varying number of candidate authors, text lengths, profile lengths and different levels of heterogeneity among the candidate authors, regarding genre, gender, and time period. The method works best when the corpora of known texts is of substantial length, when the texts being attributed are long, or when the number of candidate authors is small. If long profiles are available -- more than 60,000 words, corresponding to 150 pages of a midsize paperback book --, we demonstrated very high attribution accuracy for texts longer than a few typical novel chapters even when attributing between a large number of authors, high accuracy for texts as long as a play act or a novel chapter, and reasonable rates for short texts such as newspaper opinion pieces if the number of candidate authors is small. WANs were also shown to classify accurately the time period when a text was written, to acceptably estimate the genre of a piece, and to have some predictive power on the gender of the author. The applicability of WANs to identify multiple authors in collaborative works was also demonstrated. With regards to existing methods based on the frequency with which different function words appear in the text, we observed that WANs exceed their classification accuracy. More importantly, we showed that WANs and frequencies captured different stylometric aspects so that their combination is possible and ends up halving the error rate of existing methods.




\urlstyle{same}
\bibliographystyle{IEEEtran}
\bibliography{auth_attr_ref.bib}

\end{document}